\documentclass[letterpaper]{article} 
\usepackage{aaai24}  
\usepackage{times}  
\usepackage{helvet}  
\usepackage{courier}  
\usepackage[hyphens]{url}  
\usepackage{graphicx} 
\usepackage{natbib}  
\usepackage{caption} 
\usepackage{algpseudocode}
\usepackage[ruled,vlined]{algorithm2e}
\usepackage{amsmath}        
\usepackage{amsfonts}       
\usepackage{multirow} 
\usepackage{xcolor}%
\definecolor{lightlightgray}{gray}{0.9}
\definecolor{codebg}{rgb}{0.95,0.95,0.95}

\usepackage{newfloat}
\usepackage{listings}
\DeclareCaptionStyle{ruled}{labelfont=normalfont,labelsep=colon,strut=off} 
\title{ProAgent: Building Proactive Cooperative Agents with Large Language Models}
\author{
    Ceyao Zhang\textsuperscript{\rm 1,\rm 2}\equalcontrib\thanks{Work done when Ceyao Zhang visited Peking University.},
    Kaijie Yang\textsuperscript{\rm 3}\equalcontrib,
    Siyi Hu\textsuperscript{\rm 4}\equalcontrib,
    Zihao Wang\textsuperscript{\rm 2,5},
    Guanghe Li\textsuperscript{\rm 2},
    Yihang Sun\textsuperscript{\rm 2},\\
    Cheng Zhang\textsuperscript{\rm 2},
    Zhaowei Zhang\textsuperscript{\rm 2,5},
    Anji Liu\textsuperscript{\rm 2},
    Song-Chun Zhu\textsuperscript{\rm 5},
    Xiaojun Chang\textsuperscript{\rm 4},
    Junge Zhang\textsuperscript{\rm 3},
    Feng Yin\textsuperscript{\rm 1},
    Yitao Liang\textsuperscript{\rm 2},
    Yaodong Yang\textsuperscript{\rm 2}\thanks{Corresponding author}
}
\affiliations{
    \textsuperscript{\rm 1}SSE, The Chinese University of Hong Kong, Shenzhen\\
    \textsuperscript{\rm 2}Institute for Artificial Intelligence, Peking University\\
    \textsuperscript{\rm 3}Institute of Automation, Chinese Academy of Sciences\\
    \textsuperscript{\rm 4}ReLER, AAII, University of Technology Sydney\\
    \textsuperscript{\rm 5}National Key Laboratory of General Artificial Intelligence, Beijing Institute for General Artificial Intelligence (BIGAI)\\

    ceyaozhang2@link.cuhk.edu.cn, 
    yaodong.yang@pku.edu.cn

}

\begin{document}

\maketitle


\begin{abstract}

Building agents with adaptive behavior in cooperative tasks stands as a paramount goal in the realm of multi-agent systems. 
Current approaches to developing cooperative agents rely primarily on learning-based methods, whose policy generalization depends heavily on the diversity of teammates they interact with during the training phase.
Such reliance, however, constrains the agents' capacity for strategic adaptation when cooperating with unfamiliar teammates, which becomes a significant challenge in zero-shot coordination scenarios.
To address this challenge, we propose \textbf{ProAgent}, a novel framework that harnesses large language models (LLMs) to create \textit{pro}active \textit{agent}s capable of dynamically adapting their behavior to enhance cooperation with teammates. 
ProAgent can analyze the present state, and infer the intentions of teammates from observations. It then updates its beliefs in alignment with the teammates' subsequent actual behaviors. 
Moreover, ProAgent exhibits a high degree of modularity and interpretability, making it easily integrated into various of coordination scenarios. 
Experimental evaluations conducted within the \textit{Overcooked-AI} environment unveil the remarkable performance superiority of ProAgent, 
outperforming five methods based on self-play and population-based training when cooperating with AI agents. 
Furthermore, in partnered with human proxy models, its performance exhibits an average improvement exceeding 10\% compared to the current state-of-the-art method.
For more information about our project, please visit~\url{https://pku-proagent.github.io}.

\end{abstract}

\section{Introduction}

\begin{figure*}[t!]
    \centering
    \includegraphics[width=\textwidth]{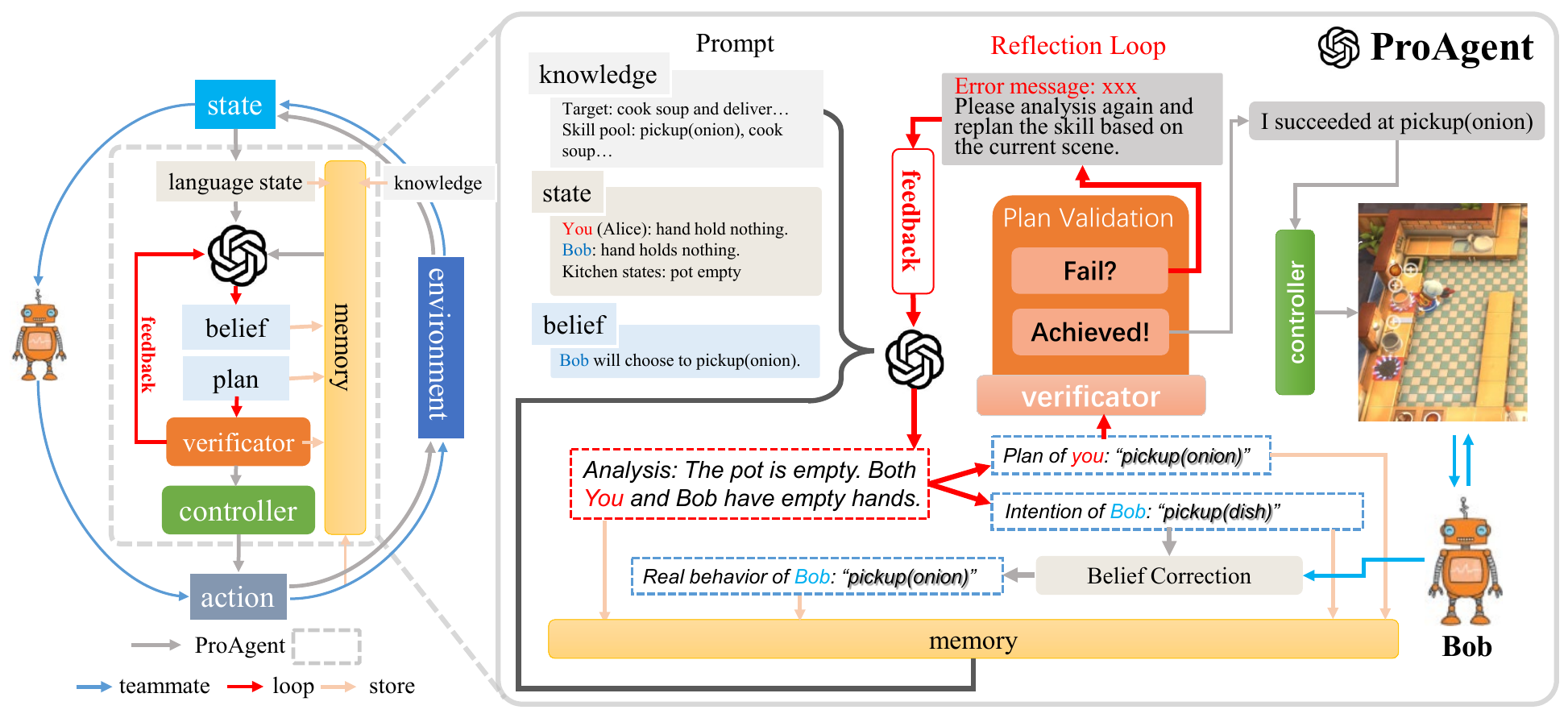}
    \caption{Overview of our proposed ProAgent framework including the coordination task workflow (\textbf{left}) and inner details of ProAgent pipeline (\textbf{right}). 
    The teammate agent's decision-making loop is formed by the blue solid arrows in the outer circle, while the decision process of the ProAgent is formed by the middle gray dotted box and the outer gray solid arrows.
    ProAgent commences its operation by translating the initial state into natural language. Then the \texttt{Planner} adeptly analyzes the provided language state in conjunction with historical information stored in the \texttt{Memory}. This analytical process allows the model to discern the \textit{intentions} of the teammate and devise a high-level \textit{skill} for the agent accordingly. 
    The belief about predicted intention will be updated through the \texttt{Belief Correction} mechanism, which involves comparing it with the subsequent actual behavior of the teammate agent. 
    As to the planned skill, the \texttt{Verificator} validates whether it can be performed under the current state.
    In case of skill failure, the \texttt{Verificator} will assess the skill's preconditions and provide a detailed explanation for the encountered issue. 
    Should the need arise, ProAgent enters into a re-plan loop, initiating a recalibration process. On the other hand, if the skill is deemed viable, the \texttt{Controller} further dissects it into several executive low-level \textit{actions}, to be executed within the environment.}
    \label{fig:ProAgent}
\end{figure*}

Large Language Models (LLMs) have rapidly emerged as powerful tools, achieving remarkable advancements across various domains, including long conversations~\citep{ouyang2022training}, reasoning~\citep{bubeck2023sparks}, and text generation~\citep{brown2020language}. These models, by leveraging a vast amount of training data, can capture and embody a significant amount of common sense knowledge. Notable LLM-based agents like SayCan~\citep{ahn2022i}, ReAct~\citep{yao2023react}, DEPS~\citep{wang2023describe}, RAP~\citep{hao2023reasoning}, Reflexion~\citep{shinn2023reflexion}, and JARVIS-1~\citep{wang2023jarvis1} have demonstrated the ability to make decisions interactively through appropriate prompts or feedback. However, these works have primarily focused on exploring the potential of LLMs as individual agents, whether in games or robotics. The untapped potential lies in investigating how LLM-based agents can effectively cooperate with other AI agents or humans. 

This research delves into the capabilities of LLMs in tackling the intricate challenges of multi-agent coordination~\citep{yang2021overview, zhang2021multi, gronauer2022multi}, particularly in the realm of policy generalization~\citep{strouse2021collaborating, zhao2023maximum, li2023cooperative, li2024tackling}. 
Current approaches~\citep{carroll2019utility,jaderberg2017population,strouse2021collaborating,zhao2023maximum,li2023cooperative, li2024tackling} to developing cooperative agents rely primarily on learning-based methods, whose policy generalization depends heavily on the diversity of teammates they interact with during the training phase.
Such reliance, however, constrains the agents' capacity for strategic adaptation when cooperating with unfamiliar teammates, which becomes a significant challenge in zero-shot coordination scenarios.
We present \textbf{ProAgent}, an innovative and adaptable framework specifically designed to excel in coordination scenarios alongside novel agents. 
ProAgent comprises four essential modules: \texttt{Planner}, \texttt{Verificator}, \texttt{Controller} and \texttt{Memory}, along with the mechanism of \texttt{Belief Correction}. 
These modules synergistically enable ProAgent to actively predict teammates' intentions and achieve adaptive cooperative reasoning and planning without the need for prior training or finetuning.
To assess the adaptive cooperative capabilities of ProAgent, we conducted performance evaluations using the well-established multi-agent coordination testing suite, \textit{Overcooked-AI}~\citep{carroll2019utility}. 
In this environment, two players must work together to maximize their score. 
The empirical findings from our evaluations reveal the following key insights:
1) ProAgent demonstrates remarkable proficiency in coordinating with various types of AI teammates across diverse scenarios.
2) ProAgent exhibits a notable preference for collaborating with rational teammates, such as the human proxy, which showcases human-like behavior and suggests its active effort to understand teammates' intentions to enhance cooperation.
These results collectively highlight the effectiveness of ProAgent as a cooperative agent across a wide range of scenarios.

In summary, our work makes three key contributions:
\textbf{Firstly}, we successfully integrate LLMs into the field of cooperative multi-agents and propose the ProAgent framework, which serves as a comprehensive guideline for leveraging the powerful reasoning and planning capabilities of LLMs in cooperative settings.
\textbf{Secondly}, we demonstrate the remarkable capability of our ProAgent to interpretably analyze the current scene, explicitly infer teammates' intentions, and dynamically adapt its behavior accordingly. This proactive nature empowers ProAgent to actively collaborate with teammates, enabling more efficient cooperative scenarios.
\textbf{Thirdly}, through a comprehensive series of experiments, we provide compelling evidence of ProAgent's superiority over other agents when engaging in cooperation with diverse types of teammates. 

\section{Related Works}

\paragraph{Reasoning and Planning with Large Language Models.}
In the realm of LLMs~\citep{huang2023reasoning, mialon2023augmented, bubeck2023sparks}, reasoning often entails decomposing intricate queries into sequential intermediate steps, referred to as Chain-of-Thought~\citep[CoT;][]{wei2022chain, kojima2022large}, to attain a final solution. Some research focuses on minimizing errors as the number of steps increases~\citep{wang2023selfconsistency}, while others explore decomposition techniques that break down complex problems into simpler subproblems~\citep{zhou2023leasttomost}. Recent endeavors have translated LLMs' reasoning capability into planning by constructing a monologue with feedback~\citep{welleck2023generating, shinn2023reflexion, paul2023refiner} to facilitate the reasoning and planning process. 
Notably, the challenge of open-ended long-term planning in the MineDojo environment~\citep{fan2022minedojo} has been addressed by utilizing LLMs as central planners~\citep{wang2023describe, wang2023jarvis1}, thereby demonstrating the extensive capabilities of LLMs-based agents in overcoming complex decision-making tasks. 
As of the camera-ready version of our paper, the application of LLM-based agents in cooperative games remains little explored.
\citet{li2023semantically} employs a centralized LLM-based planner for both two players. In contrast, our framework adopts a decentralized planning paradigm and uses one planner for one player.
While \citet{zhang2023building} also decentralized planning, their method facilitates cooperation through explicit communication. Our work, on the other hand, fosters cooperation by observing and inferring the intentions of teammates.

\paragraph{Multi-agent Coordination.} 
The goal of multi-agent coordination is to enable multiple autonomous agents to collaborate effectively towards a shared goal~\citep{rashid2018qmix, hu2020simplified, hu2021off, yu2022surprising, zhong2023heterogeneous}. However, traditional approaches have limitations in fixed task settings and struggle to handle multiple tasks or unseen scenarios. One approach to address this challenge is to enable an agent to learn multiple tasks concurrently~\citep{hu2021updet, meng2022offline, wen2022multi}. 
However, these methods may still limit the agent's cooperation ability in familiar tasks and fail to handle unseen tasks or new agent interactions. Another line of research focuses on zero-shot coordination (ZSC), utilizing Population-Based Training~\citep[PBT;][]{strouse2021collaborating, zhao2023maximum, lupu2021trajectory,lucas2022any,li2023cooperative,li2024tackling} and Theory of Mind~\citep[ToM;][]{hu2021off, wu2021too, wang2021tom2c} to facilitate adaptive policy development for coordinating with various counterparts without prior coordination experience. 
However, these ZSC methods demand significant computational resources for data collection and model optimization, and the resulting policies often lack interpretability.

\section{Method}\label{Methods}


The overview of our ProAgent framework, as is depicted in Fig.~\ref{fig:ProAgent}, involves constant interaction between agents and the environment. 
The inference pipeline of ProAgent is a hierarchical process that involves multiple interactions between the LLMs and the task at hand. 
We break down the pipeline into five key stages:

\textbf{Knowledge Library and State Grouding.} The pipeline starts with acquiring \textit{Knowledge Library} specific to the current task and transforming the raw tensor state information into \textit{Language-based State} description that the LLM can effectively comprehend.

\textbf{High-level Skill Planning.} Receiving the aligned language-based state, the LLM-based \texttt{Planner} then analyzes the current scene, infers the \textit{intentino} about the teammate agent's intentions, and plans a skill for the current agent.

\textbf{Belief Correction.} 
The belief in the teammate agent's intention is further corrected by the \texttt{Belief Correction} mechanism. 

\textbf{Skill Validation and Action Execution.} 
The selected skill will be validated by the \texttt{Verificator} and a replan is needed if the current skill fails. If a valid skill is selected, and the \texttt{Controller} module decomposes it into low-level actions, allowing ProAgent to effectively interact with the task or environment. The controller can be rule-based, or RL-based methods.

\textbf{Memory Storage.} Throughout the pipeline, all relevant information involved in the prompt, planning process, validation process, and belief correction process is stored in the \texttt{Memory} module. This accumulated knowledge helps in making informed decisions and adjusting behavior over time.

We offer a pipeline outlined in pseudocode, available in the appendix, which delineates the procedural stages of ProAgent's cooperative reasoning and planning throughout the execution of a cooperative task.
By following this pipeline, ProAgent effectively incorporates a knowledge library, performs language-based planning, and adapts its behavior through continuous interaction and memory updates. 

\subsection{Prompt Construction}

\textbf{Knowledge library}\quad The planning ability of LLMs is closely related to the prompt at the beginning, which is also the standard practice in automated planning. ProAgent is no exception, and the knowledge library should be fed into LLMs at the initial stage before the cooperation task begins. The main difficulty lies in how to build structured knowledge. In practice, we find that the best combination of knowledge library needs to be described from three perspectives, including \texttt{Task}, \texttt{Rules}, and \texttt{Demos}. We provide a template to construct the knowledge library:


\lstdefinestyle{customc}{
  belowcaptionskip=1\baselineskip,
  breaklines=true,
  frame=L,
  xleftmargin=\parindent,
  language=Bash,
  showstringspaces=false,
  basicstyle=\scriptsize,
  keywordstyle=\bfseries\color{black},
  commentstyle=\itshape\color{black},
  identifierstyle=\color{black},
  stringstyle=\color{orange},
}

\lstset{escapechar=@,
style=customc, 
numbers=none, 
backgroundcolor=\color{codebg},
mathescape=true,
}

\begin{lstlisting}
### Task: 
- The task requires two players player0 and player1 to work together as a team ...
- To get the points, the team needs to ...
...
### Rules:
In this task, each player can ONLY perform the following skills: [skill 1], [skill 2], ...
def skill_1(obj):
    [function detail]        
def skill_2(obj, obj):  
    [function detail]
...
Suppose you are an assistant who is proficient in the task. Your goal is to control player0 and cooperate with player1 who is controlled by a certain strategy to get a high score and should follow:
- [Rule 1]. 
- [Rule 2].
...
- Based on the current scene, you need to achieve
- [Target 1].
- [Target 2]. 
...
Your response should be in the following format:
- Analysis:[your analysis of the current scene]
- Plan for Player 0: [one skill in [skill 1, skill 2...]]
...
### Demos:
Scene 0: [Player0 state 0]. [Player1 state 0]. [Other task information]
Analysis: Both Player0 and Player1 are [State description]. I guess two players will [Some skill].
[Target 1]: [Some skill].
[Target 2]: [Some skill].
...
Scene 39: [Player0 state 39]. [Player1 state 39]. [Other task information]
Analysis: Player0 is [State description]. Player1 is [State description]. I guess ...
...
\end{lstlisting}

% ...

\texttt{Task} is for LLMs to understand the objective of the task and information about other cooperative agents. \texttt{Rules} is designed to regulate the planning pattern of the LLM, defining which skills are legal and which are not. In \texttt{Rules}, we can also enforce the format of LLMs' responses to follow the CoT: output analysis and then plan according to the analysis instead of directly outputting a plan. \texttt{Demos} is an optional component of the three. Its main functionality is to provide real cases for LLMs to strengthen their memories and behave following the regulations set by \texttt{Rules}. Normally, \texttt{Demos} should contain a scene description followed by an analysis and the desired behavior, such as the selected skill. With these three parts, LLMs can understand the task and what is expected of them in the subsequent planning and reasoning stages.

\noindent\textbf{Grounding tensor state to language-based state}\quad
To facilitate interaction between LLMs and the environment, it is essential to establish a bridge between the original symbolic state provided by the environment and the language-based state for LLMs. In most scenarios, the raw state is not directly applicable to LLMs' usage. Hence, finding an effective alignment between the original symbolic state and the language-based state is crucial to enhancing LLMs' accurate understanding of the current situation. To illustrate this, we present a simplified example based on the \textit{Overcooked-AI}  environment, demonstrating how the state can be transformed into language within our ProAgent framework.
With the knowledge library and initial state information prepared, ProAgent is equipped to tackle the cooperative task alongside its teammates. This marks the transition to the subsequent stage, where ProAgent engages in reasoning and planning, progressing step by step to achieve its objectives. Here is an illustrative instance of the alignment between state representation and natural language:



\begin{lstlisting}
### Original State 
        ___________________________________
        |X       X       P       X       X|       
        |                                 |
        |O       $\leftarrow{}$0o     $\leftarrow{}$1o               O|  
        |                                 |
        |X                               X|       
        |                                 |
        |X       D       X       S       X|    
        -----------------------------------
### Language State (Layout)
Above is the layout of the kitchen: onion dispenser at (0, 1), onion dispenser at (4, 1), dish dispenser at (1, 3), pot at (2, 0), serving loc at (3, 3).
### Language State (Task state)
State: Player 0 holds one onion. Player 1 holds one onion. Kitchen states: Pot (2, 0) is empty. 
\end{lstlisting}

\subsection{Cooperative Reasoning and Planning}

ProAgent is a specialized system tailored for cooperative tasks, where information from teammate agents plays a pivotal role in the coordination process. Existing works mainly utilize information in two ways: firstly, through explicit incorporation, involving communication and exchange of information before decision-making; secondly, through implicit modeling of teammate agents to facilitate cooperative learning. Each approach comes with its own set of advantages and disadvantages concerning cooperative reasoning and planning:
The integration of teammate information can be achieved efficiently by sending teammate agent information to LLMs. However, this approach may jeopardize the overall generalization of ProAgent's reasoning capabilities.
On the other hand, modeling the teammate agent offers a more flexible approach, while the modeling process is inherently unstable as the teammate agent's strategy may continuously evolve, demanding additional resources for maintenance.

In order to strike a balance between the generalization ability of built agents and the efficiency of incorporating teammate information, particularly for LLMs that possess excellent reasoning capabilities but face challenges in fine-tuning or learning extra belief modules, ProAgent introduces three core components along with a cooperative reasoning and planning mechanism. The three modules encompass:
1) The \texttt{Memory} module, which stores information about task trajectory and general knowledge in the task domain.
2) The \texttt{Verificator} module, consisting of one component for skill failure analysis and another for transforming skills into atomic actions.
3) The \texttt{Controller} module, dedicated to the transformation of skills into atomic actions.
To further align the LLMs' belief regarding the teammate agent's intentions with actual behavior, and thereby continually enhance prediction accuracy, ProAgent implements the \texttt{Belief Correction} mechanism. This process effectively strengthens the LLMs' beliefs, leading to improved cooperative reasoning and planning.


\subsubsection{Memory Module: Leveraging History for Cooperative Behavior}

In ProAgent, the \texttt{Memory} module plays a crucial role in supporting information storage and retrieval processes. It consists of two components: \texttt{Knowledge Library} and \texttt{Trajectory}.
The \texttt{Knowledge Library} acts as a persistent repository, retaining a comprehensive record of the task, including its layout, rules, and demonstrations throughout gameplay sessions.
On the other hand, the \texttt{Trajectory} component serves as a temporary buffer with a fixed length, following a \textit{First-In, First-Out} (FIFO) approach. It stores essential information, such as the latest \texttt{Language-based State}, \texttt{Analysis}, \texttt{Belief} of teammates' intentions, and the \texttt{Skill} used, while discarding the most outdated data. When needed, only specific parts of the \texttt{Memory} are retrieved, depending on the chosen strategy, such as the \texttt{recent-K} strategy\footnote{only retrieve the $K$ most recent trajectories.}. This strategy focuses on the immediate context, facilitating efficient decision-making and planning during ongoing interactions.
Overall, the \texttt{Memory} module significantly enhances ProAgent's capacity to access pertinent information and cooperate efficiently with teammate agents. By leveraging past experiences and learning from historical data, the \texttt{Memory} module empowers ProAgent to make informed decisions during cooperation tasks.

\subsubsection{Planner Module: Reasoning with Chain of Thought}

With the history information and current state description ready, ProAgent utilizes the strong reasoning ability of LLMs to make decisions in the current situation. The \texttt{Planner} module, which follows the Chain of Thought (CoT) approach commonly used in LLMs' reasoning and planning work~\citep{yao2023react, hao2023reasoning, shinn2023reflexion}. Instead of directly outputting a plan, the \texttt{Planner} module makes the final decision step by step. The provided information is first thoroughly analyzed, and the intention of the teammate agent's plan for the current step is predicted. Based on this \texttt{Analysis} and the \texttt{Belief} about the teammate agent, LLMs formulate a plan that ensures it is the most reasonable and effective strategy for the given situation.
In the experiment part, we conduct an ablation study to assess how this design enhances ProAgent's performance in a cooperative scenario. 

\subsubsection{Verificator Module: Analyzing Skill Failures With Multi-rounds Prompts}\label{subsec:verificator}

In the cooperative setting, the \texttt{Verificator} module plays a crucial role in scrutinizing and identifying any unreasonable or flawed planning generated by the LLMs. Its primary function involves analyzing the underlying reasons for these inadequacies and providing valuable insights and suggestions for improvement. In the ProAgent framework, this process entails conducting a thorough investigation through multiple rounds of prompt and response between the agent and the LLMs.

To illustrate this process, we present an example based on \textit{Overcooked-AI} , where we employ a three-round prompt and response approach, including \texttt{Preconditions Check}, \texttt{Double-check}, and \texttt{Error Conclusion}. It's important to note that the number of rounds or the specific interaction style is not restricted, and the core idea behind the \texttt{Verificator} module remains focused on decomposing how to replan for the current agent when receiving negative feedback from external environments by checking and determination.

\textbf{Preconditions Check}:
The \texttt{Preconditions Check} involves signaling the LLMs if the current plan is illegal due to internal checks before its actual execution. A robust internal checking mechanism can prevent failures when the LLMs haven't fully understood the consequences of their chosen skill under the current state. In the \textit{Overcooked-AI} example, we design the condition check prompt by leveraging both the current scene and the failed skill as inputs. We employ a trigger prompt to enable the LLMs to individually verify each precondition of the skill and pinpoint the specific one that led to the failure. To aid in solving multi-step reasoning problems, prompting techniques like CoT are also adopted. An instance of the trigger prompt in Overcooked-AI could be: "\textit{Analysis of why I cannot execute this skill in the current scene step by step.}"
The preconditions of each skill can be expressed either in natural language or in pseudo-code form, which can be more effective as proposed in previous works~\citep{liang2023code, singh2023progprompt}. 

\subsubsection{Belief Correction: Rectifying Belief on Teammate Agents}

The \texttt{Belief Correction} mechanism plays a pivotal role in rectifying any incorrect beliefs during cooperation. ProAgent makes predictions about their teammates' future behavior and stores relevant analyses in their memory. In subsequent steps, ProAgent verifies the accuracy of their predictions and corrects any erroneous beliefs. Specifically, if the observed behavior of the teammate agent deviates from the assumed intentions recorded in \texttt{Memory}, the \texttt{Belief Correction} mechanism can take two approaches: 
1) Replace the predicted intention with the actual behavior of the teammate.
2) Provide an annotation alongside the original prediction to flag it as incorrect.
The replacement method enforces ProAgent to learn from ground truth, while the annotation method allows ProAgent to reason about the cause of the wrong belief, thereby avoiding similar mistakes in the future.
Additionally, the replan loop within the \texttt{Verificator} module serves as an indirect method for rectifying beliefs. With each query to the LLMs, ProAgent outputs new intentions on their teammate agent, which contributes to improving the accuracy of their predictions. This iterative process allows ProAgent to refine their beliefs over time and enhance their ability to make accurate predictions about their teammate's intentions. In summary, the \texttt{Belief Correction} mechanism ensures that ProAgent maintains accurate and up-to-date information about their teammate agent's real behavior. By referencing the \texttt{Belief} part of \texttt{Memory} before making decisions, ProAgent continually improves the accuracy of their beliefs regarding their teammate's future behavior.

\begin{table*}[t]
\centering
\renewcommand{\arraystretch}{1.1} 
\resizebox{1.0\textwidth}{!}{
\begin{tabular}{p{3.5cm}cccccc} 
\hline
\multirow{2}{*}{\textbf{Layout}} 
& \multicolumn{5}{c}{\textbf{Baseline AI Agents}} & \multirow{2}{*}{\textbf{ProAgent} (ours)} \\
\cline{2-6}
&  SP & PBT & FCP & MEP & COLE \\
\hline
\multirow{2}{*}{\textbf{Cramped Room}} 
& $168.5 \pm 15.2$ & $178.8 \pm 16.5$ & $196.3 \pm 16.8$ & $185 \pm 15$ & $163.8 \pm 24.1$ & $\mathbf{197.3} \pm 6.1$ \\
& $172.8 \pm 16.1$ & $179.8 \pm 26.8$ & $\mathbf{196} \pm 11.9$ & $178.2 \pm 15.6$ & $169.2 \pm 16.8$ & $194.2 \pm 10.5$ \\
\hline
\multirow{2}{*}{\textbf{Asymmetric Advantages}} 
& $183.3 \pm 27.5$ & $182.2 \pm 27.9$ & $185.7 \pm 22.7$ & $155.7 \pm 63.9$ & $201.3 \pm 34.5$ & $\mathbf{228.7} \pm 23$ \\
& $177.8 \pm 24.6$ & $152.3 \pm 64.5$ & $167.8 \pm 21.3$ & $184 \pm 41.8$ & $165.5 \pm 33.3$ & $\mathbf{229.8} \pm 21.9$ \\
\hline
\multirow{2}{*}{\textbf{Coordination Ring}} 
& $122 \pm 17.2$ & $141.3 \pm 28$ & $148.8 \pm 19.4$ & $167.2 \pm 22.4$ & $168.8 \pm 26.1$ & $\mathbf{175.3} \pm 29$ \\
& $133.3 \pm 23.7$ & $141.3 \pm 27.5$ & $145.7 \pm 17.1$ & $159.3 \pm 25.3$ & $158.3 \pm 27.1$ & $\mathbf{183} \pm 31.7$ \\
\hline
\multirow{2}{*}{\textbf{Forced Coordination}} 
& $6.7 \pm 6.7$ & $15.3 \pm 17.1$ & $44.7 \pm 36.4$ & $23.3 \pm 19.8$ & $24 \pm 21.8$ & $\mathbf{49.7} \pm 33.1$ \\
& $30.2 \pm 21.9$ & $\mathbf{61.7} \pm 46$ & $32.2 \pm 30.2$ & $39.3 \pm 16.9$ & $57.3 \pm 36.4$ & $31 \pm 33.9$ \\
\hline
\multirow{2}{*}{\textbf{Counter Circuit}} 
& $64.7 \pm 45.8$ & $64.7 \pm 45.9$ & $58.3 \pm 37.5$ & $74.3 \pm 39.1$ & $95.5 \pm 25.2$ & $\mathbf{126.3} \pm 32.3$ \\
& $60.7 \pm 40.8$ & $54.3 \pm 49.1$ & $60 \pm 38.3$ & $81.5 \pm 27.5$ & $100.8 \pm 31.1$ & $\mathbf{128.5} \pm 28.1$ \\
\hline
\end{tabular}
}
\caption{Performance for all AI agent pairs. Each column represents the average reward and standard error of one algorithm playing with all others. For each layout, the first row represents the scenario where the agent takes the role of Player 0, and the AI partner takes the role of Player 1. The second row depicts the vice-versa scenario. The best results for each layout are highlighted in bold.}
\label{table:ai-ave-bar}
\end{table*}

\subsubsection{Controller Module: Grounding High-Level Skills to Low-Level Actions}

Based on the modules and mechanisms discussed above, ProAgent effectively engages in cooperative reasoning and plans a high-level skill. 
However, it is worth noting that there is a gap between the skill space and the environment's action space.
Therefore, we also need a \texttt{Controller} module which is imperative, aiming to convert language-based skills into low-level actions that can be executed in the environment. 
Although this transformation process is closely tied to the specific task at hand, making the \texttt{Controller} module highly flexible, it necessitates the establishment of fixed rules capable of decomposing the skill into multiple steps of low-level actions and providing a feedback signal to the reasoning component once the action is fully executed. 
The controller can be a rule-based path search algorithm or a policy trained by language-grounded reinforcement learning~\citep{hanjie2021grounding, ding2023entity, hu2023language, du2023guiding} methods.
Considering that the controller is not our main concern, we choose the built-in controller in the Overcooked-AI environment based on Best-First Search and a better controller can definitely reach better performance.
An example of how 
the skill \texttt{fill\_dish\_with\_soup()} is executed and completed in three timesteps can be found in the appendix.
% ...

 
% ...


\section{Experiments}\label{sec:exp}

\subsection{Experimental Settings}\label{subsec:settings}

Following previous works on cooperative AI and human-AI cooperation, we choose Overcooked-AI as our test environment, in which two agents swiftly prepare and serve soups by placing up to three ingredients in a pot, cooking the soup, filling the soup with the dish, and delivering the soup. 
Agents must dynamically allocate tasks and cooperate effectively. 
Five classical layouts are used: \textit{Cramped Room}, \textit{Asymmetric Advantages}, \textit{Forced Coordination}, \textit{Coordination Ring}, and \textit{Counter Circuit}. 
A detailed description of each layout can be found in the appendix.

Our primary concern behind this work is how well the agents developed so far based on ZSC methods can cooperate with diverse teammates, ranging from different AI agents to humans.
In previous works on Overcooked-AI, the cooperative performance of an agent is often evaluated with two held-out populations: self-play (SP) agent and human proxy model. 
We conduct a comparative analysis between our proposed ProAgent and five alternatives prevalent in the field including SP~\citep{tesauro1994td, carroll2019utility}, PBT~\citep{jaderberg2017population}, FCP~\citep{strouse2021collaborating}, MEP~\citep{zhao2023maximum}, and COLE~\citep{li2023cooperative, li2024tackling}. 
We combined the above six algorithms in pairs to construct 36 pairs.
For example, we choose the SP algorithm as player 0 and the PBT algorithm as player 1, and these two algorithms can form an agent pair (SP, PBT). 
Since the two players are not all homogeneous, we will also form a (PBT, SP) algorithm pair. 
For each algorithm pair, we ran five episodes and collected the mean and standard variation of the episode returns.
Besides, we also select the human proxy model proposed by~\citep{carroll2019utility} to test the agent's ability to cooperate with humans.

\subsection{Collaborating with AI Agents}\label{subsubsec:CAI}

\paragraph{Quantitative Results}
Table~\ref{table:ai-ave-bar} illustrates the average performance of SP, PBT, FCP, MEP, COLE, and ProAgent when paired with all the others. For each layout, the first row represents the scenario where the agent takes the role of Player 0, and the AI partner takes the role of Player 1. The second row depicts the vice-versa scenario.
The results indicate that ProAgent outperforms the baselines in all layouts when acting as Playe 0. Taking the role of Player 1, ProAgent only slightly underperforms FCP in cramped room layout and loses to PBT in forced coordination layout.
We will examine this failure further in the appendix.
In previous studies, it is rare to compare different AI agent combinations with each other, and our experimental results also reveal that none of the other ZSC methods is consistently better than other methods.
Considering that ProAgent requires no specific training with distinct teammates and in distinct layouts, it presents a stronger adaptive ability than the other AI agents. 
These results show our LLM-based agent is a better cooperator.

\begin{figure*}[t]
    \centering
    \includegraphics[width=\textwidth]{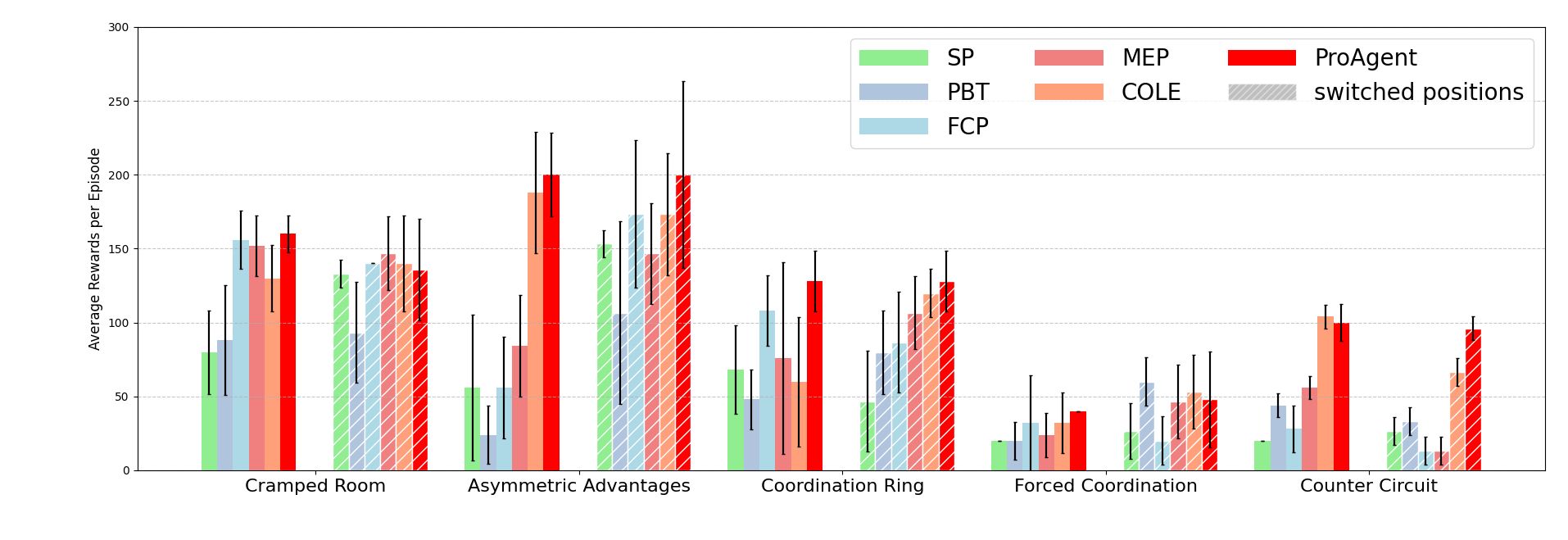}
    \caption{Performance with human proxy partners. In each layout, the reward bar represents the average performance of one algorithm collaborating with the unseen human proxy partners over 400 timesteps on five BC models, and the error lines represent the standard error. The hashed bars indicate the rewards obtained where the starting positions are switched.}
    \label{fig:bc-ave-bar}
\end{figure*}

\paragraph{Qualitative Results} 
To gain deeper insights into the fundamental components of effective cooperation, we perform a qualitative examination of our ProAgent's behaviors exhibited during our experiments, leading us to identify several cooperative behaviors.

\textit{ProAgent excels in making strategic plans.} 
For example, when pot one is cooking and pot two lacks an onion, we observed that ProAgent would prioritize putting one onion into pot two.
After this, the agent will fetch the plate. At the same time, cooking can be completed in the first pot, and this agent with a plate can directly fill the plate with soup. This process is very effective.
Besides, after making a failure plan, ProAgent can promptly recognize this failure, and make a new and often better plan.

\textit{ProAgent demonstrates a remarkable capacity to dynamically adjust low-level actions while executing high-level plans.} 
For instance, when ProAgent intends to deposit an onion into a pot, it's underlying \texttt{Controller} identifies a blocked path caused by its teammate. Swiftly, the \texttt{Controller} will identify an alternative interconnected route, skillfully bypassing any potential obstructions. 
This adaptive strategy enables ProAgent to discover unhindered pathways. 
Moreover, when \texttt{Planner} has no clear goal, the \texttt{Controller} will move randomly.
This dynamic operation helps ProAgent to break the deadlock caused by other AI agents due to conventions formed during the training phase.

\subsection{Collaborating with Humans}\label{subsubsec:CHuman} 

Apart from cooperation with AI agents, our concern also involves the generalization to human partners. Due to the limitation of collecting human interaction data, we follow the previous work~\citep{carroll2019utility} that uses a behavior cloning (BC) model trained on human data as a proxy of humans.
Fig.~\ref{fig:bc-ave-bar} presents the average cumulative rewards achieved for 400 timesteps by ProAgent when engaged in collaboration with BC. 
The reported outcomes encompass both the mean value and standard error across five distinct BC models. 
Analysis of the experimental findings reveals that across the five environments, ProAgent outperforms the baseline in four environments, exhibiting particularly noteworthy superiority when functioning as Player 0 in the context of \textit{Forced Coordination}. 
Notably, the positioning discrepancy between the left and right starting positions had a negligible impact on ProAgent's performance. 
However, this difference led to substantial performance disparities among the baselines, particularly in asymmetric layouts, where the cumulative rewards achieved by all baselines were superior in the left position compared to the right position, consistent with the findings in COLE~\citep{li2023cooperative, li2024tackling}.



\subsection{Discussion}\label{subsec:analysis}

\paragraph{Does analysis and belief help in better planning?}
To gauge the influence of \textit{analysis} and \textit{belief} on the accuracy and efficiency of decisions made by the \texttt{Planner} Module, we conducted an ablation study within the context of the \textit{Cramped Room} layout.
The experiment considered three distinct conditions and their respective scores were: 1) 204 for with both \textit{analysis} and \textit{belief}, 2) 184 without \textit{belief}, and 3) 100 for no \textit{analysis} and \textit{belief}, and making a skill plan directly. 
We believe that the significance of analysis in the \texttt{Planner} Module lies in its provision of in-context for final planning just as CoT will improve the effect of reasoning.
Additionally, inferring teammate intentions provides further improvements.

\paragraph{Is Verificator effective in feedback-based reasoning?}
Upon removing the \texttt{Verificator} Module and allowing ProAgent to engage in planning without feedback, we computed success rates over 100 steps. Notably, the success rate dropped significantly to 20\%, underscoring the critical role of our \texttt{Verificator} Module in furnishing feedback when the \texttt{Planner} Module generates inaccurate plans.


\section{Conclusion}\label{sec:conclution}
In this work, we propose \textbf{ProAgent}, a proactive LLM-based agent framework, with the primary objective of addressing the multi-agent coordination predicament. 
By leveraging the inherent faculties of LLMs encompassing common sense comprehension and language-centric task understanding, coupled with explicit mechanisms for reasoning and planning, ProAgent demonstrates remarkable performance within various coordination scenarios. 
Experiments on cooperating with both AI agents and human proxies in the Overcooked-AI demonstrate the effectiveness of ProAgent over state-of-the-art methods. 
Moreover, ProAgent's reasoning and planning are based on natural language, which is interpretable and friendly to humans.
These encouraging results pave the way for further advancements in both cooperative multi-agent and human-compatible AI systems built upon LLMs.

\section*{Acknowledgement}
This work is sponsored by the National Natural Science Foundation of China (62376013), by the Basic Research Project No. HZQB-KCZYZ-2021067 of Hetao Shenzhen-HK S$\&$T Cooperation Zone, Beijing Municipal Science $\&$ Technology Commission (Z231100007423015), by the Shenzhen Outstanding Talents Training Fund 202002, by the Guangdong Research Projects No. 2017ZT07X152 and No. 2019CX01X104, by the Guangdong Provincial Key Laboratory of Future Networks of Intelligence (Grant No. 2022B1212010001), by the Shenzhen Key Laboratory of Big Data and Artificial Intelligence (Grant No. ZDSYS201707251409055), by the NSFC under Grant No. 62271433, and by Shenzhen Science and Technology Program under Grant No. JCYJ20220530143806016 and No. RCJC20210609104448114.

\bibliography{ref}

\begin{thebibliography}{48}
\providecommand{\natexlab}[1]{#1}

\bibitem[{Ahn et~al.(2022)Ahn, Brohan, Brown, Chebotar, Cortes, David, Finn, Fu, Gopalakrishnan, Hausman, Herzog, Ho, Hsu, Ibarz, Ichter, Irpan, Jang, Ruano, Jeffrey, Jesmonth, Joshi, Julian, Kalashnikov, Kuang, Lee, Levine, Lu, Luu, Parada, Pastor, Quiambao, Rao, Rettinghouse, Reyes, Sermanet, Sievers, Tan, Toshev, Vanhoucke, Xia, Xiao, Xu, Xu, Yan, and Zeng}]{ahn2022i}
Ahn, M.; Brohan, A.; Brown, N.; Chebotar, Y.; Cortes, O.; David, B.; Finn, C.; Fu, C.; Gopalakrishnan, K.; Hausman, K.; Herzog, A.; Ho, D.; Hsu, J.; Ibarz, J.; Ichter, B.; Irpan, A.; Jang, E.; Ruano, R.~J.; Jeffrey, K.; Jesmonth, S.; Joshi, N.~J.; Julian, R.; Kalashnikov, D.; Kuang, Y.; Lee, K.-H.; Levine, S.; Lu, Y.; Luu, L.; Parada, C.; Pastor, P.; Quiambao, J.; Rao, K.; Rettinghouse, J.; Reyes, D.; Sermanet, P.; Sievers, N.; Tan, C.; Toshev, A.; Vanhoucke, V.; Xia, F.; Xiao, T.; Xu, P.; Xu, S.; Yan, M.; and Zeng, A. 2022.
\newblock Do As I Can, Not As I Say: Grounding Language in Robotic Affordances.
\newblock arXiv:2204.01691.

\bibitem[{Brown et~al.(2020)Brown, Mann, Ryder, Subbiah, Kaplan, Dhariwal, Neelakantan, Shyam, Sastry, Askell et~al.}]{brown2020language}
Brown, T.; Mann, B.; Ryder, N.; Subbiah, M.; Kaplan, J.~D.; Dhariwal, P.; Neelakantan, A.; Shyam, P.; Sastry, G.; Askell, A.; et~al. 2020.
\newblock Language Models are Few-Shot Learners.
\newblock In \emph{Advances in neural information processing systems}, volume~33, 1877--1901.

\bibitem[{Bubeck et~al.(2023)Bubeck, Chandrasekaran, Eldan, Gehrke, Horvitz, Kamar, Lee, Lee, Li, Lundberg, Nori, Palangi, Ribeiro, and Zhang}]{bubeck2023sparks}
Bubeck, S.; Chandrasekaran, V.; Eldan, R.; Gehrke, J.; Horvitz, E.; Kamar, E.; Lee, P.; Lee, Y.~T.; Li, Y.; Lundberg, S.; Nori, H.; Palangi, H.; Ribeiro, M.~T.; and Zhang, Y. 2023.
\newblock Sparks of Artificial General Intelligence: Early Experiments with GPT-4.
\newblock arXiv:2303.12712.

\bibitem[{Carroll et~al.(2019)Carroll, Shah, Ho, Griffiths, Seshia, Abbeel, and Dragan}]{carroll2019utility}
Carroll, M.; Shah, R.; Ho, M.~K.; Griffiths, T.; Seshia, S.; Abbeel, P.; and Dragan, A. 2019.
\newblock On the Utility of Learning about Humans for Human-AI Coordination.
\newblock In \emph{Advances in neural information processing systems}, volume~32.

\bibitem[{Ding et~al.(2023)Ding, Zhang, Yue, Wang, Huang, and Lu}]{ding2023entity}
Ding, Z.; Zhang, W.; Yue, J.; Wang, X.; Huang, T.; and Lu, Z. 2023.
\newblock Entity Divider with Language Grounding in Multi-Agent Reinforcement Learning.
\newblock In \emph{International Conference on Machine Learning}, 8103--8119. PMLR.

\bibitem[{Du et~al.(2023)Du, Watkins, Wang, Colas, Darrell, Abbeel, Gupta, and Andreas}]{du2023guiding}
Du, Y.; Watkins, O.; Wang, Z.; Colas, C.; Darrell, T.; Abbeel, P.; Gupta, A.; and Andreas, J. 2023.
\newblock Guiding Pretraining in Reinforcement Learning with Large Language Models.
\newblock arXiv:2302.06692.

\bibitem[{Fan et~al.(2022)Fan, Wang, Jiang, Mandlekar, Yang, Zhu, Tang, Huang, Zhu, and Anandkumar}]{fan2022minedojo}
Fan, L.; Wang, G.; Jiang, Y.; Mandlekar, A.; Yang, Y.; Zhu, H.; Tang, A.; Huang, D.-A.; Zhu, Y.; and Anandkumar, A. 2022.
\newblock MineDojo: Building Open-Ended Embodied Agents with Internet-Scale Knowledge.
\newblock In \emph{NIPS Processing Systems Datasets and Benchmarks Track}.

\bibitem[{Gronauer and Diepold(2022)}]{gronauer2022multi}
Gronauer, S.; and Diepold, K. 2022.
\newblock Multi-Agent Deep Reinforcement Learning: A survey.
\newblock \emph{Artificial Intelligence Review}, 1--49.

\bibitem[{Hanjie, Zhong, and Narasimhan(2021)}]{hanjie2021grounding}
Hanjie, A.~W.; Zhong, V.~Y.; and Narasimhan, K. 2021.
\newblock Grounding Language to Entities and Dynamics for Generalization in Reinforcement Learning.
\newblock In \emph{International Conference on Machine Learning}, 4051--4062. PMLR.

\bibitem[{Hao et~al.(2023)Hao, Gu, Ma, Hong, Wang, Wang, and Hu}]{hao2023reasoning}
Hao, S.; Gu, Y.; Ma, H.; Hong, J.~J.; Wang, Z.; Wang, D.~Z.; and Hu, Z. 2023.
\newblock Reasoning with Language Model is Planning with World Model.
\newblock arXiv:2305.14992.

\bibitem[{Hu and Foerster(2020)}]{hu2020simplified}
Hu, H.; and Foerster, J.~N. 2020.
\newblock Simplified Action Decoder for Deep Multi-Agent Reinforcement Learning.
\newblock In \emph{International Conference on Learning Representations}.

\bibitem[{Hu et~al.(2021{\natexlab{a}})Hu, Lerer, Cui, Pineda, Brown, and Foerster}]{hu2021off}
Hu, H.; Lerer, A.; Cui, B.; Pineda, L.; Brown, N.; and Foerster, J. 2021{\natexlab{a}}.
\newblock Off-Belief Learning.
\newblock In \emph{International Conference on Machine Learning}, 4369--4379. PMLR.

\bibitem[{Hu and Sadigh(2023)}]{hu2023language}
Hu, H.; and Sadigh, D. 2023.
\newblock Language Instructed Reinforcement Learning for Human-AI Coordination.
\newblock In \emph{Proceedings of the 40th International Conference on Machine Learning}. PMLR.

\bibitem[{Hu et~al.(2021{\natexlab{b}})Hu, Zhu, Chang, and Liang}]{hu2021updet}
Hu, S.; Zhu, F.; Chang, X.; and Liang, X. 2021{\natexlab{b}}.
\newblock UPDeT: Universal Multi-agent Reinforcement Learning via Policy Decoupling with Transformers.
\newblock arXiv:2101.08001.

\bibitem[{Huang and Chang(2023)}]{huang2023reasoning}
Huang, J.; and Chang, K. C.-C. 2023.
\newblock Towards Reasoning in Large Language Models: A Survey.
\newblock arXiv:2212.10403.

\bibitem[{Jaderberg et~al.(2017)Jaderberg, Dalibard, Osindero, Czarnecki, Donahue, Razavi, Vinyals, Green, Dunning, Simonyan, Fernando, and Kavukcuoglu}]{jaderberg2017population}
Jaderberg, M.; Dalibard, V.; Osindero, S.; Czarnecki, W.~M.; Donahue, J.; Razavi, A.; Vinyals, O.; Green, T.; Dunning, I.; Simonyan, K.; Fernando, C.; and Kavukcuoglu, K. 2017.
\newblock Population Based Training of Neural Networks.
\newblock arXiv:1711.09846.

\bibitem[{Kojima et~al.(2022)Kojima, Gu, Reid, Matsuo, and Iwasawa}]{kojima2022large}
Kojima, T.; Gu, S.~S.; Reid, M.; Matsuo, Y.; and Iwasawa, Y. 2022.
\newblock Large Language Models are Zero-Shot Reasoners.
\newblock In \emph{Advances in neural information processing systems}, volume~35, 22199--22213.

\bibitem[{Li et~al.(2023{\natexlab{a}})Li, Qiao, Wang, Wang, Jin, and Zha}]{li2023semantically}
Li, W.; Qiao, D.; Wang, B.; Wang, X.; Jin, B.; and Zha, H. 2023{\natexlab{a}}.
\newblock Semantically Aligned Task Decomposition in Multi-Agent Reinforcement Learning.
\newblock arXiv:2305.10865.

\bibitem[{Li et~al.(2023{\natexlab{b}})Li, Zhang, Sun, Du, Wen, Wang, and Pan}]{li2023cooperative}
Li, Y.; Zhang, S.; Sun, J.; Du, Y.; Wen, Y.; Wang, X.; and Pan, W. 2023{\natexlab{b}}.
\newblock Cooperative Open-ended Learning Framework for Zero-shot Coordination.
\newblock In \emph{Proceedings of the 40th International Conference on Machine Learning}. PMLR.

\bibitem[{Li et~al.(2024)Li, Zhang, Sun, Zhang, Du, Wen, Wang, and Pan}]{li2024tackling}
Li, Y.; Zhang, S.; Sun, J.; Zhang, W.; Du, Y.; Wen, Y.; Wang, X.; and Pan, W. 2024.
\newblock Tackling Cooperative Incompatibility for Zero-Shot Human-AI Coordination.
\newblock arXiv:2306.03034.

\bibitem[{Liang et~al.(2023)Liang, Huang, Xia, Xu, Hausman, Ichter, Florence, and Zeng}]{liang2023code}
Liang, J.; Huang, W.; Xia, F.; Xu, P.; Hausman, K.; Ichter, B.; Florence, P.; and Zeng, A. 2023.
\newblock Code as Policies: Language Model Programs for Embodied Control.
\newblock In \emph{2023 IEEE International Conference on Robotics and Automation (ICRA)}, 9493--9500. IEEE.

\bibitem[{Lucas and Allen(2022)}]{lucas2022any}
Lucas, K.; and Allen, R.~E. 2022.
\newblock Any-Play: An Intrinsic Augmentation for Zero-Shot Coordination.
\newblock In \emph{International Foundation for Autonomous Agents and Multiagent Systems}, 853–861.

\bibitem[{Lupu et~al.(2021)Lupu, Cui, Hu, and Foerster}]{lupu2021trajectory}
Lupu, A.; Cui, B.; Hu, H.; and Foerster, J. 2021.
\newblock Trajectory Diversity for Zero-Shot Coordination.
\newblock In \emph{International conference on machine learning}, 7204--7213. PMLR.

\bibitem[{Meng et~al.(2022)Meng, Wen, Yang, Le, Li, Zhang, Wen, Zhang, Wang, and Xu}]{meng2022offline}
Meng, L.; Wen, M.; Yang, Y.; Le, C.; Li, X.; Zhang, W.; Wen, Y.; Zhang, H.; Wang, J.; and Xu, B. 2022.
\newblock Offline Pre-trained Multi-Agent Decision Transformer: One Big Sequence Model Tackles All SMAC Tasks.
\newblock arXiv:2112.02845.

\bibitem[{Mialon et~al.(2023)Mialon, Dessì, Lomeli, Nalmpantis, Pasunuru, Raileanu, Rozière, Schick, Dwivedi-Yu, Celikyilmaz, Grave, LeCun, and Scialom}]{mialon2023augmented}
Mialon, G.; Dessì, R.; Lomeli, M.; Nalmpantis, C.; Pasunuru, R.; Raileanu, R.; Rozière, B.; Schick, T.; Dwivedi-Yu, J.; Celikyilmaz, A.; Grave, E.; LeCun, Y.; and Scialom, T. 2023.
\newblock Augmented Language Models: a Survey.
\newblock arXiv:2302.07842.

\bibitem[{Ouyang et~al.(2022)Ouyang, Wu, Jiang, Almeida, Wainwright, Mishkin, Zhang, Agarwal, Slama, Ray, Schulman, Hilton, Kelton, Miller, Simens, Askell, Welinder, Christiano, Leike, and Lowe}]{ouyang2022training}
Ouyang, L.; Wu, J.; Jiang, X.; Almeida, D.; Wainwright, C.; Mishkin, P.; Zhang, C.; Agarwal, S.; Slama, K.; Ray, A.; Schulman, J.; Hilton, J.; Kelton, F.; Miller, L.; Simens, M.; Askell, A.; Welinder, P.; Christiano, P.~F.; Leike, J.; and Lowe, R. 2022.
\newblock Training Language Models to Follow Instructions with Human Feedback.
\newblock In \emph{Advances in Neural Information Processing Systems}, volume~35, 27730--27744.

\bibitem[{Paul et~al.(2023)Paul, Ismayilzada, Peyrard, Borges, Bosselut, West, and Faltings}]{paul2023refiner}
Paul, D.; Ismayilzada, M.; Peyrard, M.; Borges, B.; Bosselut, A.; West, R.; and Faltings, B. 2023.
\newblock REFINER: Reasoning Feedback on Intermediate Representations.
\newblock arXiv:2304.01904.

\bibitem[{Rashid et~al.(2018)Rashid, Samvelyan, Schroeder, Farquhar, Foerster, and Whiteson}]{rashid2018qmix}
Rashid, T.; Samvelyan, M.; Schroeder, C.; Farquhar, G.; Foerster, J.; and Whiteson, S. 2018.
\newblock QMIX: Monotonic Value Function Factorisation for Deep Multi-Agent Reinforcement Learning.
\newblock In \emph{International Conference on Machine Learning}, 4295--4304. PMLR.

\bibitem[{Shinn et~al.(2023)Shinn, Cassano, Gopinath, Narasimhan, and Yao}]{shinn2023reflexion}
Shinn, N.; Cassano, F.; Gopinath, A.; Narasimhan, K.~R.; and Yao, S. 2023.
\newblock Reflexion: Language Agents with Verbal Reinforcement Learning.
\newblock In \emph{Thirty-seventh Conference on Neural Information Processing Systems}.

\bibitem[{Singh et~al.(2023)Singh, Blukis, Mousavian, Goyal, Xu, Tremblay, Fox, Thomason, and Garg}]{singh2023progprompt}
Singh, I.; Blukis, V.; Mousavian, A.; Goyal, A.; Xu, D.; Tremblay, J.; Fox, D.; Thomason, J.; and Garg, A. 2023.
\newblock ProgPrompt: Generating Situated Robot Task Plans using Large Language Models.
\newblock In \emph{2023 IEEE International Conference on Robotics and Automation (ICRA)}, 11523--11530. IEEE.

\bibitem[{Strouse et~al.(2021)Strouse, McKee, Botvinick, Hughes, and Everett}]{strouse2021collaborating}
Strouse, D.; McKee, K.; Botvinick, M.; Hughes, E.; and Everett, R. 2021.
\newblock Collaborating with Humans without Human Data.
\newblock In \emph{Advances in Neural Information Processing Systems}, volume~34, 14502--14515.

\bibitem[{Tesauro(1994)}]{tesauro1994td}
Tesauro, G. 1994.
\newblock TD-Gammon, a self-teaching backgammon program, achieves master-level play.
\newblock \emph{Neural computation}, 6(2): 215--219.

\bibitem[{Wang et~al.(2023{\natexlab{a}})Wang, Wei, Schuurmans, Le, Chi, Narang, Chowdhery, and Zhou}]{wang2023selfconsistency}
Wang, X.; Wei, J.; Schuurmans, D.; Le, Q.~V.; Chi, E.~H.; Narang, S.; Chowdhery, A.; and Zhou, D. 2023{\natexlab{a}}.
\newblock Self-Consistency Improves Chain of Thought Reasoning in Language Models.
\newblock In \emph{The Eleventh International Conference on Learning Representations}.

\bibitem[{Wang et~al.(2021)Wang, Zhong, Xu, and Wang}]{wang2021tom2c}
Wang, Y.; Zhong, F.; Xu, J.; and Wang, Y. 2021.
\newblock ToM2C: Target-oriented Multi-agent Communication and Cooperation with Theory of Mind.
\newblock In \emph{International Conference on Learning Representations}.

\bibitem[{Wang et~al.(2023{\natexlab{b}})Wang, Cai, Chen, Liu, Ma, and Liang}]{wang2023describe}
Wang, Z.; Cai, S.; Chen, G.; Liu, A.; Ma, X.; and Liang, Y. 2023{\natexlab{b}}.
\newblock Describe, Explain, Plan and Select: Interactive Planning with LLMs Enables Open-World Multi-Task Agents.
\newblock In \emph{Thirty-seventh Conference on Neural Information Processing Systems}.

\bibitem[{Wang et~al.(2023{\natexlab{c}})Wang, Cai, Liu, Jin, Hou, Zhang, Lin, He, Zheng, Yang, Ma, and Liang}]{wang2023jarvis1}
Wang, Z.; Cai, S.; Liu, A.; Jin, Y.; Hou, J.; Zhang, B.; Lin, H.; He, Z.; Zheng, Z.; Yang, Y.; Ma, X.; and Liang, Y. 2023{\natexlab{c}}.
\newblock JARVIS-1: Open-World Multi-Task Agents with Memory-Augmented Multimodal Language Models.
\newblock arXiv:2311.05997.

\bibitem[{Wei et~al.(2022)Wei, Wang, Schuurmans, Bosma, brian ichter, Xia, Chi, Le, and Zhou}]{wei2022chain}
Wei, J.; Wang, X.; Schuurmans, D.; Bosma, M.; brian ichter; Xia, F.; Chi, E.~H.; Le, Q.~V.; and Zhou, D. 2022.
\newblock Chain of Thought Prompting Elicits Reasoning in Large Language Models.
\newblock In \emph{Advances in Neural Information Processing Systems}, volume~35, 24824--24837.

\bibitem[{Welleck et~al.(2023)Welleck, Lu, West, Brahman, Shen, Khashabi, and Choi}]{welleck2023generating}
Welleck, S.; Lu, X.; West, P.; Brahman, F.; Shen, T.; Khashabi, D.; and Choi, Y. 2023.
\newblock Generating Sequences by Learning to Self-Correct.
\newblock In \emph{The Eleventh International Conference on Learning Representations}.

\bibitem[{Wen et~al.(2022)Wen, Kuba, Lin, Zhang, Wen, Wang, and Yang}]{wen2022multi}
Wen, M.; Kuba, J.; Lin, R.; Zhang, W.; Wen, Y.; Wang, J.; and Yang, Y. 2022.
\newblock Multi-Agent Reinforcement Learning is a Sequence Modeling Problem.
\newblock In \emph{Advances in Neural Information Processing Systems}, volume~35, 16509--16521.

\bibitem[{Wu et~al.(2021)Wu, Wang, Evans, Tenenbaum, Parkes, and Kleiman-Weiner}]{wu2021too}
Wu, S.~A.; Wang, R.~E.; Evans, J.~A.; Tenenbaum, J.~B.; Parkes, D.~C.; and Kleiman-Weiner, M. 2021.
\newblock Too Many Cooks: Bayesian Inference for Coordinating Multi-Agent Collaboration.
\newblock \emph{Topics in Cognitive Science}, 13(2): 414--432.

\bibitem[{Yang and Wang(2021)}]{yang2021overview}
Yang, Y.; and Wang, J. 2021.
\newblock An Overview of Multi-Agent Reinforcement Learning from Game Theoretical Perspective.
\newblock arXiv:2011.00583.

\bibitem[{Yao et~al.(2023)Yao, Zhao, Yu, Du, Shafran, Narasimhan, and Cao}]{yao2023react}
Yao, S.; Zhao, J.; Yu, D.; Du, N.; Shafran, I.; Narasimhan, K.~R.; and Cao, Y. 2023.
\newblock ReAct: Synergizing Reasoning and Acting in Language Models.
\newblock In \emph{The Eleventh International Conference on Learning Representations}.

\bibitem[{Yu et~al.(2022)Yu, Velu, Vinitsky, Gao, Wang, Bayen, and Wu}]{yu2022surprising}
Yu, C.; Velu, A.; Vinitsky, E.; Gao, J.; Wang, Y.; Bayen, A.; and Wu, Y. 2022.
\newblock The Surprising Effectiveness of PPO in Cooperative Multi-Agent Games.
\newblock In \emph{Advances in Neural Information Processing Systems}, volume~35, 24611--24624.

\bibitem[{Zhang et~al.(2023)Zhang, Du, Shan, Zhou, Du, Tenenbaum, Shu, and Gan}]{zhang2023building}
Zhang, H.; Du, W.; Shan, J.; Zhou, Q.; Du, Y.; Tenenbaum, J.~B.; Shu, T.; and Gan, C. 2023.
\newblock Building Cooperative Embodied Agents Modularly with Large Language Models.
\newblock arXiv:2307.02485.

\bibitem[{Zhang, Yang, and Ba{\c{s}}ar(2021)}]{zhang2021multi}
Zhang, K.; Yang, Z.; and Ba{\c{s}}ar, T. 2021.
\newblock Multi-Agent Reinforcement Learning: A Selective Overview of Theories and Algorithms.
\newblock \emph{Handbook of reinforcement learning and control}, 321--384.

\bibitem[{Zhao et~al.(2023)Zhao, Song, Yuan, Hu, Gao, Wu, Sun, and Yang}]{zhao2023maximum}
Zhao, R.; Song, J.; Yuan, Y.; Hu, H.; Gao, Y.; Wu, Y.; Sun, Z.; and Yang, W. 2023.
\newblock Maximum Entropy Population Based Training for Zero-Shot Human-AI Coordination.
\newblock In \emph{Proceedings of the AAAI Conference on Artificial Intelligence}, 5, 6145--6153.

\bibitem[{Zhong et~al.(2023)Zhong, Kuba, Feng, Hu, Ji, and Yang}]{zhong2023heterogeneous}
Zhong, Y.; Kuba, J.~G.; Feng, X.; Hu, S.; Ji, J.; and Yang, Y. 2023.
\newblock Heterogeneous-Agent Reinforcement Learning.
\newblock arXiv:2304.09870.

\bibitem[{Zhou et~al.(2023)Zhou, Sch{\"a}rli, Hou, Wei, Scales, Wang, Schuurmans, Cui, Bousquet, Le, and Chi}]{zhou2023leasttomost}
Zhou, D.; Sch{\"a}rli, N.; Hou, L.; Wei, J.; Scales, N.; Wang, X.; Schuurmans, D.; Cui, C.; Bousquet, O.; Le, Q.~V.; and Chi, E.~H. 2023.
\newblock Least-to-Most Prompting Enables Complex Reasoning in Large Language Models.
\newblock In \emph{The Eleventh International Conference on Learning Representations}.

\end{thebibliography}

\end{document}